\documentclass[11pt]{article}
\usepackage{common}
\usepackage{graphicx}
\usepackage{hyperref}
\graphicspath{ {images/} }
\usepackage{listings}
\usepackage{amsmath,amsthm,amssymb,amsfonts}
\usepackage{enumitem}

\hypersetup{
    colorlinks=true,
    linkcolor=blue,
    filecolor=magenta,      
    urlcolor=blue,
    citecolor=blue,
}
\title{Reinforcement Learning for Stock Transactions
}
\author{Ziyi (Queena) Zhou\thanks{Harvard John A. Paulson School of Engineering and Applied Sciences}\and Nicholas Stern\footnotemark[1]\and Julien Laasri\footnotemark[1]}
\date{December 19, 2018}
\begin{document}
\maketitle{}

\section{Introduction}

Much research has been done to analyze the stock market. After all, if one can determine a pattern in the chaotic frenzy of transactions, then they could make a hefty profit from capitalizing on these insights. As such, the goal of our project was to apply reinforcement learning (RL) to determine the best time to buy a stock within a given time frame. With only a few adjustments, our model can be extended to identify the best time to sell a stock as well. In order to use the format of free, real-world data to train the model, we define our own Markov Decision Process (MDP) problem. These two papers \cite{lee2001stock} \cite{orderbook} helped us in formulating the state space and the reward system of our MDP problem. We train a series of agents using Q-Learning, Q-Learning with linear function approximation, and deep Q-Learning. In addition, we try to predict the stock prices using machine learning regression and classification models. We then compare our agents to see if they converge on a policy, and if so, which one learned the best policy to maximize profit on the stock market.

\section{Background and Related Work}

In this section we provide the necessary financial background to understand our results, and some related work that uses RL to learn the stock market. Observable stock price exhibits an almost continuous pattern, under the constraint that each transaction is made at a discrete point in time. We cannot observe the price at all times of the day because the stock exchanges are only open during a select period of time. During the open days of the stock exchanges, each stock has an associated opening, closing, high and low price for that day. Opening and closing prices are merely "snapshots" of the process, whereas high and low prices are constantly monitored to be updated \cite{ohlc}. To maximize their profits, people would want to buy stocks at a relatively low price and sell stocks at a relatively high price. However, stock prices are highly random, so having the ability to predict stock trends is extremely useful before making trading decisions.

Many studies use RL techniques to model the stock market. An RL system \cite{lee2001stock} has been developed to predict the stock price using temporal difference (TD) learning. It maps each stock price trend into a state and directly updates the value of the expected future cumulative reward for each state. To approximate these state values, it builds a multi-layer neural network. Another RL system \cite{orderbook} takes millisecond time-scale limit order data from NASDAQ to train the trading execution. It uses a modified Q-learning algorithm which updates the expected cost associated with taking each action instead of Q-value. It shows that RL approaches are well-suited for optimized execution and can result in substantial improvements in performance. 

Researchers have also implemented supervised learning to predict the stock price. This paper \cite{ding2015deep} builds a deep convolutional neural network to predict stock price movements based on events. This paper \cite{tsai2009stock} combines artificial neural network with decision trees to forecast stock price and is able to achieve 77\% accuracy. 

\section{Problem Specification}

As our overarching goal is to make money from identifying implicit stock market trends, we train an agent that can recognize the best time to buy a stock within a given time frame. This meshes well with the daily data we obtain from \href{https://finance.yahoo.com/}{Yahoo! Finance}, as we have price information stretching back to each companies' inception.\footnote{We originally planned to have our agent dynamically buy/sell specific amounts of stock at millisecond time scales. Unfortunately, and understandably, the data to accomplish this were locked behind substantial pay walls.}

\subsection{Data}

Taking a closer look at the data, we observe that each day has four prices associated with it. These prices signify the open, high, low, and close price of the stock on that day. Throughout each day, stocks will typically fluctuate on the order of a few cents to a dollar, not changing too drastically on any single day, but adding up over time. We scraped the stock price history for four tech companies: Apple, Amazon, Microsoft, and Google. We choose to gather data from similar companies to eradicate any cross-industry variance that might affect our agent comparisons down the line. 

\begin{figure}[!h]
\centering
\includegraphics[width=17cm]{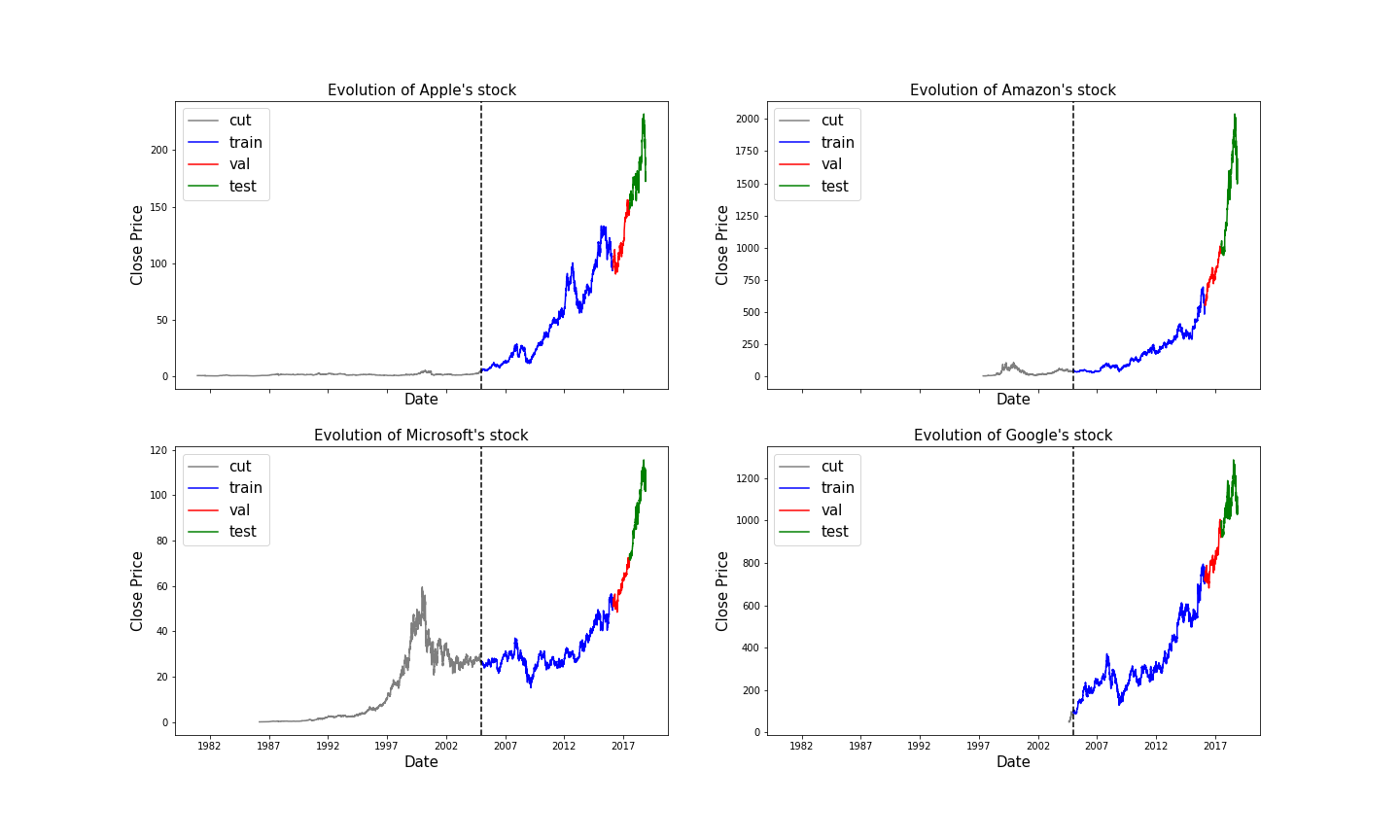}
\caption{A visualization of the stock history data for the four companies we examined. The grey section is the part we cut out, the blue section is the training data, the red is the validation data, and the green is the testing data. From this plot one can see that the data we are training our agents on is similar across the four companies, and also across the train/val/test split.}
\label{data_cut}
\end{figure}

Furthermore, we subset the stock history to only consider data more recent than 2005 in order to ensure that:
\begin{enumerate}
    \item The training, validation, and test sets within a company have similar trends.
    \item Agents would be trained on data from the same overall time frame, since Google does not have publicly logged stock prices before 2005. 
\end{enumerate}
Figure \ref{data_cut} shows how this cut on the data improved the similarity between company trends, as well as the train, validation, test splits for the data, which were done using 80/10/10 divisions.  

\subsection{MDP Framework}
\begin{figure}
    \centering
    \includegraphics[width=8cm]{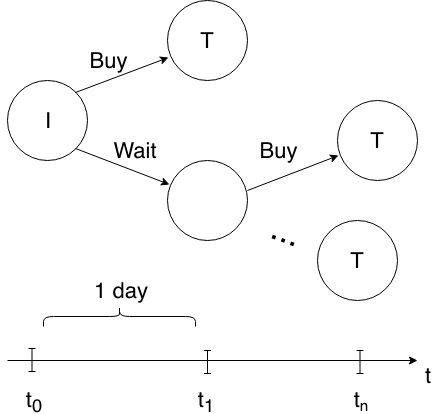}
    \caption{A drawing of our Markov Decision Process. The initial state is marked with an I, while terminal states are marked with T's. A decision is made on each day, shown on the timeline at the bottom. The timeline represents a single time window.}
    \label{mdp}
\end{figure}

For our project, we choose to implement a binary agent with two possible actions: "buy" and "wait." The decision to omit the action to sell allows us to define an intuitive problem framework that is easily generalizable to a symmetric agent that can either "sell" or "wait." We define our problem in terms of the following steps:

\begin{itemize}
    \item Split the stock price data into subsets of $w$ days called \textit{time windows}.
    \item Each day corresponds to a state, in which the agent must make an action (buy or wait). Note that the agent MUST make a purchase within the time window, in order to ensure the agent does not sit idly forever. Also note that the states can include previous days' trends, dubbed the \textit{history}, to better inform the agent's decision.
    \item Once the agent chooses to buy, we skip the remaining days in the current time window and transition to the first state of the next time window.
    \item The reward for the agent's purchase is based on the amount of money it made/lost within that time window. There is no reward for waiting, but a penalty for waiting until the last day when the agent will be forced to make a purchase.
\end{itemize}
These series of steps are represented as Markov Decision Process in Figure \ref{mdp}. 

\section{Approach}

\subsection{Agents}
The following sections outline the specifics of how we develop our agents within the context of the problem description. We work on four agents in total. The first is a baseline agent that implements some simple logic to set the standard. The next is an $\epsilon$-greedy Q-learning agent with a reduced state space. And the last two are $\epsilon$-greedy agents that use approximate Q-learning to predict the Q-values for a continuous state space, one using a linear function, and the other using a neural network.

\subsubsection{Baseline}
We define a baseline agent that is inspired from the submit and leave (S\&L) agent described by Nevmyvaka et al., 2006 \cite{orderbook}. The agent has only one constant parameter $d$. Given a time window, the agent buys the stock only if the price decreases to \$$d$ less than the initial price. As all agents have to buy during the time window (by how we defined our MDP), we also force this agent to buy at the last time frame if it does not buy the stock before that.

\subsubsection{Exact Q-Learning agent}\label{mvt_def}
Q-learning is the process of iteratively updating the value of a (state, action) pair with a function of the rewards obtained from each pass through the tuple. This means that a proper Q-learning framework must reduce the state space so that each (state, action) pair is visited multiple times, otherwise the Q-values will not converge. In order to create a Q-learning agent that fits within the framework of our problem, we abstract away daily prices into daily \textit{movements} instead. These movements signifies the directionality of the price trend within a particular day, which is the sign of the difference between the close price and the open price. Each state contains $h$ previous movements, corresponding to the desired number of previous days' trends we wish the agent to consider, as well as the trend of the current day. It is necessary to keep $h$, the history size, quite small so as not to inflate the size of the state space and risk convergence. 

Because we lack specific price information within each state, we decide to use a simple reward function that gives a constant positive reward $r$, for when the agent buys before a price increase the next day (meaning it bought at a low), and consequentially a negative reward $-r$ for when the agent buys before the price dropped the next day (meaning it bought at a high). Unfortunately we are not able to make comparisons to an initial price, and therefore to encourage the agent to buy, instead of forcing the agent to buy within a time window, we give a slight negative reward for choosing to wait $-c$. However, when evaluating the agents together, we take the actions of the Q-learning agent and evaluate how it would have performed within the context of time windows, even though it is trained to make day to day decisions from historical trends. An example of state representation, the action space, and reward function is summarized below:

\begin{align*}
    \text{State} &= (\text{up}, \text{down}, \text{up}) \\
    \text{Actions} &= (\text{buy}, \text{wait}) \\
    \text{Reward} &= \begin{cases} r, & \mbox{if buy then increase} \\ 
    -r, & \mbox{if buy then decrease} \\ -c, & \mbox{if wait}\end{cases}
\end{align*}
Note that in the example state representation above, $h = 2$, while the exact significance is: two days ago the price went up, one day ago the price went back down, and today the price went up again. 

\subsubsection{Approximate Q-Learning: Linear}
The benefit of using approximate Q-learning in our problem framework is that we are allowed to have a continuous state space. This is essential because it is very rare that the vector of prices for any given stock on any given day will be the same again in the future. The idea behind approximate Q-learning is that we can predict the Q-values of states given a set of features, $f(s,a)$, which are functions of states $s$ and actions $a$. Linear function approximation learns weights, $w_i$, associated with these features, and produces a linear combination of these features as the prediction of the Q-value. In our case, we choose to have each feature be a binary indicator for a (state, action) pair. In other words, we have $s\cdot a$ features, each with an associated weight. This process is formalized through the following equations:

\begin{align*}
    Q(s,a) &= w_1f_1(s,a) + w_2f_2(s,a) + \dots + w_nf_n(s,a) \\
    w_i &\leftarrow w_i + \alpha\big[r + \gamma\max_{a'}{Q(s', a')}\big]f_i(s,a)
\end{align*}
where $\alpha$ corresponds to the learning rate and $\gamma$ signifies the discount factor. 

As before, we include historical information in the states, although this time we are able to include specific prices. Therefore, the state representation is modified to include all four descriptive prices for any given day included in the state. The reward is set to be the negative of the difference in price between the day the agent decides to buy $p_0$ and the initial day in the time frame $p_i$. The negative sign ensures that positive rewards are allotted for buying at lower prices. An example of state representation, action space and reward function is provided below: 

\begin{align*}
    \text{State} &= (\vec{p}_{-3}, \vec{p}_{-2}, \vec{p}_{-1}, \vec{p}_{0}) \\
    \text{Actions} &= (\text{buy}, \text{wait}) \\
    \text{Reward} &= \begin{cases} -(p_0 - p_i), & \mbox{if buy} \\
    0, & \mbox{if wait}\end{cases}
\end{align*}
where $\vec{p}_{-3}$ stands for the vector of prices 3 days ago, $\vec{p}_{-2}$ is the vector of prices 2 days ago, etc.

\subsubsection{Approximate Q-Learning: Deep Neural Network}
As in Q-Learning with linear function approximation, deep Q-Learning tries to solve the issue where the state space is continuous and massive. Unlike the linear approximation, deep Q-Learning utilizes a neural network structure to store and approximate the Q-value for a given (state, action) pair. For example, we design a neural network that takes in inputs $s$ and $a$, and outputs $Q(s,a)$. We use the squared Bellman error to be our loss function: $$(Q(s,a) - (r+\gamma \max_{a'}{Q(s',a')}))^2.$$ Because our action set has a small size, we decide to use two separate neural networks for each action we have. For example, the neural network for action 'buy' will take $s$ as input and output $Q(s, buy)$. Our deep Q-Learning algorithm comes with some additional hyperparameters: 
\begin{itemize}
    \item The number of hidden layers
    \item The number of hidden units in each layer
\end{itemize}
We tune these two hyperparameters on a hold-out validation set. 

\subsection{Price prediction}

Predicting tomorrow's stock price could be used as a feature in the state space of our agents to develop better policies. We use a simple 80/20 train/test split on Apple stock data. Alltogether, we endeavor to try two different approaches:
\begin{enumerate}
    \item Regression formulation: Predicting tomorrow's close price based on today's information using a linear regression.
    \item Classification formulation: Predict tomorrow's \textit{movement} (up or down) as defined in Section \ref{mvt_def} based on 3 previous days' information, using XGBoost\cite{xgboost}, LightGBM\cite{lightgbm} and Logistic Regression.
\end{enumerate}

\section{Experimental results}
\subsection{Price prediction}
\subsubsection{Regression formulation}
\begin{figure}[!h]
\centering
\includegraphics[width=10cm]{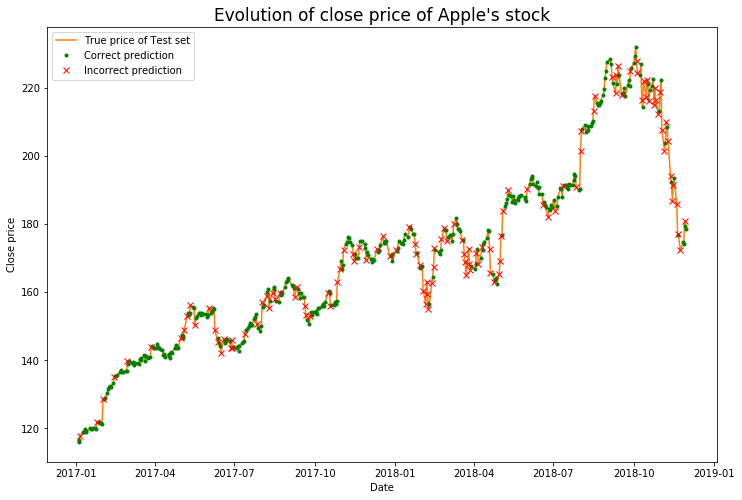}
\caption{Accuracy of our predictions on the test set for the regression formulation of the problem.}
\label{prediction_accuracy}
\end{figure}

For the regression formulation, we define a correct prediction as predicting the close price within an error of at most 2\%. Doing so, our model is correct 75.52\% of the time on the test set. This seems quite good, but with a deeper investigation we can expose a flaw in the predictions of our model.
As shown in Figure \ref{prediction_accuracy}, we only manage to obtain correct predictions when the price doesn't change much. In the turning points of the price, our predictions are not correct.

\begin{figure}[!h]
\centering
\includegraphics[width=10cm]{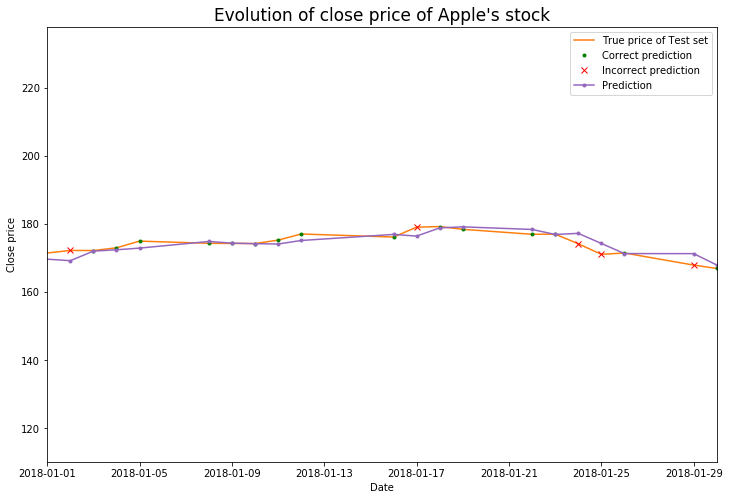}
\caption{Accuracy of our predictions on the test set for the regression formulation of the problem.}
\label{prediction_zoom}
\end{figure}

Furthermore, if we zoom in on our predictions on the test set as shown in Figure \ref{prediction_zoom}, we see that our prediction for the next day is just the price of the current day. We decide to implement a baseline that would predict tomorrow's day by simply predicting it as being the same as today's close price. This baseline has a similar performance on the test set, with a prediction accuracy of 76.56\%. Unfortunately, this type of prediction would not provide our agents with information they don't already have. Next, we turn to classification. 

\subsubsection{Classification formulation}
Predicting tomorrow's \textit{movement} based on the three last days' information leads to the results in Table \ref{results_classification}.

\begin{table}[h!]
\centering
\begin{tabular}{c|c|c} 
 \hline
Classification algorithm & Train accuracy & Test accuracy \\
 \hline
 Logistic Regression & 0.5098 & 0.4917 \\ 
 XGBoost & 0.7352 & 0.4772 \\
 LightGBM & 0.9275 & 0.4793 \\
 \hline
\end{tabular}
\caption{Price prediction results with classification formulation.}
\label{results_classification}
\end{table}

As you can see from this table, the predictions on the test set are close to being random. The gradient boosting methods, which would be liable to pick up complex nonlinearities, seem to overfit drastically to the training data. The fluctuations of the market seem to be too random for our classification algorithms to capture any real trends that could help us predict even just the movement of the price of future days (perhaps understandably so, otherwise we would have no need for reinforcement learning in the first place). We thus decide to abandon the research on these price prediction algorithms as they did not seem to provide us with valuable information as to the future fluctuations of the price as the market.

\subsection{Agents}

\subsubsection{Evaluation strategy}

We evaluate our agents by running them on unseen test data without updating the Q-values or the weights they used to predict the Q-values. We then measure the amount of money they were able to make from purchasing at the time they did within the window. We compute this in the same way we computed the reward for the approximate Q-learning agents, by taking the negative of the difference in close price on the day they decide to buy vs. the first day of the time window. This difference of close price is the score that we give to our agent for that time window. If the agent never decides to buy for that time window, we force it to buy at the last timestep. We use the validation set to tune the hyperparameters of our agents, and tested all of them on the unseen test set. Our final score is the average score of how our agent performs on all time windows of the unseen test data. We denote this to be the average \textit{profit} of our agents.

\begin{figure}[!h]
\centering
\includegraphics[width=17cm]{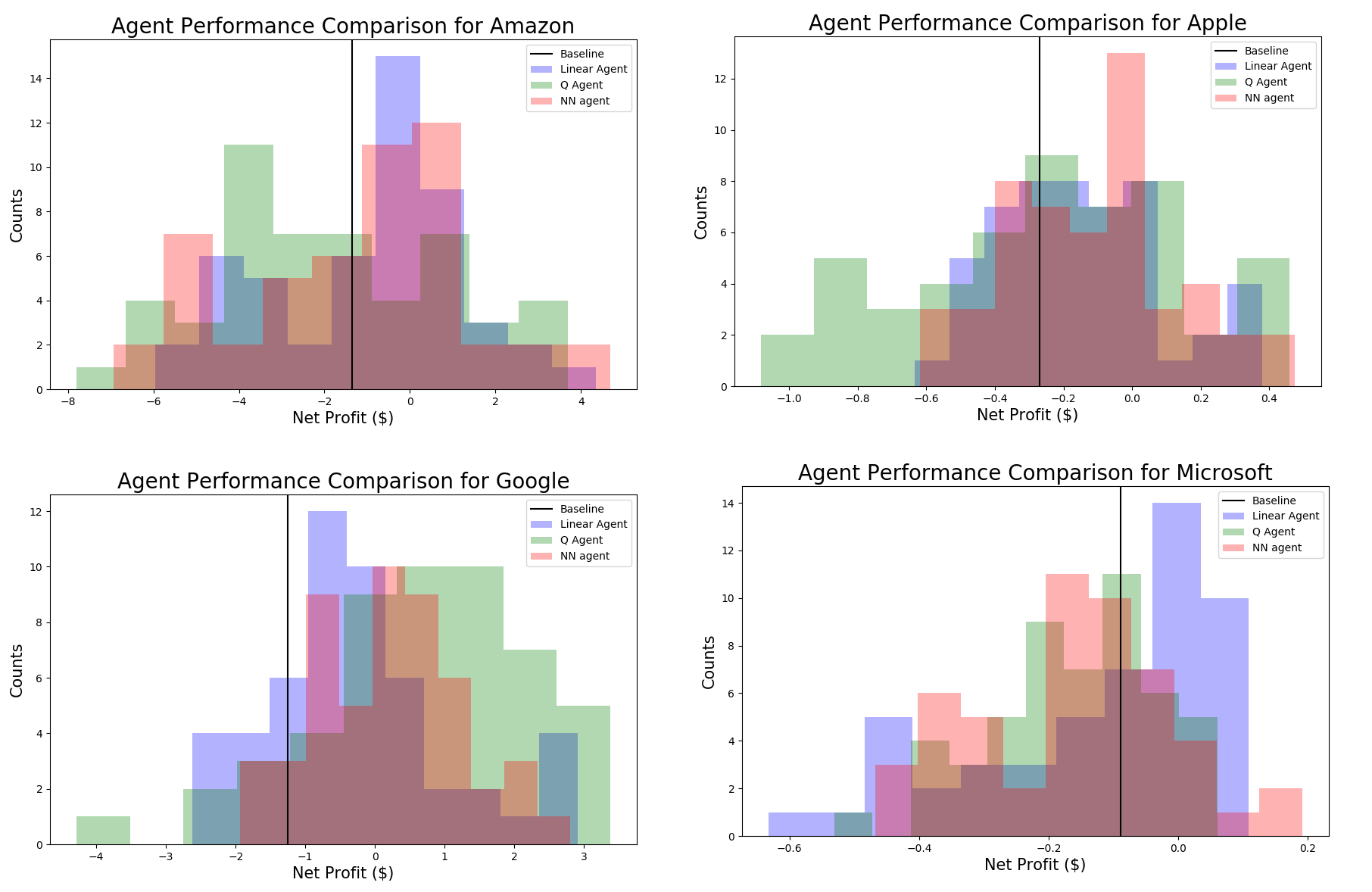}
\caption{Histograms of the results for each company.}
\label{perfs_histograms}
\end{figure}

We realize that there is a lot of variation in the validation scores for models that had the exact same parameters, and only a small epsilon value. As such, it becomes apparent that our agents are not able to converge on a policy. Therefore, we decide to run our scoring regime 51 times for each agent to generate distributions of their average profit for each company as shown in figure \ref{perfs_histograms}.

\subsubsection{Results}

The confidence intervals are Student confidence intervals, our values being considered to be gaussians with unknown standard deviations.

The performance of each of our agents for each company can be found in tables \ref{perf_apple}, \ref{perf_amazon}, \ref{perf_microsoft}, and \ref{perf_google}.

\begin{table}[h!]
\centering
\begin{tabular}{c|c|c|c}
\hline
Agent              & Average Profit & 95\% CI & Profit Stdev \\
\hline
Baseline           & -0.2704       & {[}-0.2704;-0.2704{]}              & 0.0                      \\
Q-Learning         & -0.2482       & {[}-0.3556;-0.1407{]}              & 0.3819                   \\
Approximate Linear & -0.1529       & {[}-0.2213;-0.0845{]}              & 0.2432                   \\
Deep Q-Learning    & -0.1143       & {[}-0.1851;-0.0435{]}              & 0.2516                      
\end{tabular}
\caption{Performance of our agents on Apple}
\label{perf_apple}
\end{table}

\begin{table}[h!]
\centering
\begin{tabular}{c|c|c|c}
\hline
Agent              & Average Profit & 95\% CI & Profit Stdev \\
\hline
Baseline           & -1.3497       & {[}-1.3497;-1.3497{]}              & 0.0                      \\
Q-Learning         & -1.9154       & {[}-2.6890;-1.1417{]}              & 2.7501                   \\
Approximate Linear & -0.9481       & {[}-1.5752;-0.3210{]}              & 2.2292                   \\
Deep Q-Learning    & -1.2765       & {[}-2.0159;-0.5371{]}              & 2.6282                  
\end{tabular}
\caption{Performance of our agents on Amazon}
\label{perf_amazon}
\end{table}

\begin{table}[h!]
\centering
\begin{tabular}{c|c|c|c}
\hline
Agent              & Average Profit & 95\% CI & Profit Stdev \\
\hline
Baseline           & -0.089        & {[}-0.0890;-0.0890{]}              & 0.0                      \\
Q-Learning         & -0.1602       & {[}-0.1960;-0.1243{]}              & 0.1275                   \\
Approximate Linear & -0.126        & {[}-0.1785;-0.0734{]}              & 0.1868                   \\
Deep Q-Learning    & -0.1593       & {[}-0.2025;-0.1161{]}              & 0.1537                  
\end{tabular}
\caption{Performance of our agents on Microsoft}
\label{perf_microsoft}
\end{table}

\begin{table}[h!]
\centering
\begin{tabular}{c|c|c|c}
\hline
Agent              & Average Profit & 95\% CI & Profit Stdev \\
\hline
Baseline           & -1.2574       & {[}-1.2574;-1.2574{]}              & 0.0                      \\
Q-Learning         & 0.662         & {[}0.2177;1.1063{]}                & 1.5793                   \\
Approximate Linear & -0.2434       & {[}-0.6277;0.1409{]}               & 1.3662                   \\
Deep Q-Learning    & 0.2075        & {[}-0.0939;0.5090{]}               & 1.0717                  
\end{tabular}
\caption{Performance of our agents on Google}
\label{perf_google}
\end{table}

\section{Discussion}

The first thing to note from the histograms is that there is a significant spread to the performance of each agent. This is visual confirmation that the agents' policies have not converged, and therefore they perform differently according to the random fluctuations provided by their $\epsilon$ value. This could be due to several reasons. Firstly, it is possible, as hinted by the price prediction effort, that the stock trends are so random that the agent cannot identify any implicit trends in the data. Secondly, we know that there are many external factors that impact stock trends, such as poor publicity, global shortages, company policies, etc. Our agents are ignorant to all of these trends, so what seems like random behavior to them, may be explained by external events. Finally, it is possible that we do not have enough data for our agents to converge on a consistent policy. Our data is relatively low-frequency (daily), as opposed to some of the RL agents we read about that used very high-frequency data (millisecond timescales), of which there is much more to be found. 

That being said, we have done our best to get a general sense for how the agents performed from our averaging scheme. Looking at the tabular data above, we can see that for each company, the agent that performs the best varies. For Apple, the Deep Learning agent performed the best on average, followed by the linear agent, the Q-learning agent, and then the baseline. This trend follows our expectation, since the Deep Learning agent is most likely to capture nonlinear implicit trends, whereas the others are simpler models, and the order of performs is aligned with the order of simplicity. The company that exhibits characteristics most contrary to our expectations turned out to be Microsoft, as not only is the standard deviation of the agents' performances much lower, but the order is nearly reversed, in that the baseline model performs the best on average, as opposed to the RL agents. This could mean that there are subtle effects that confuse the agents, whereas the simpler model abstracts away the subtle impacts and gives a better general sense for where the trends are going. 

This highlights the strengths and weaknesses of each approach. The strengths of the complex models involve highlighting nonlinear, complex trends within the stock history. This is beneficial when these trends follow a hard to find pattern that repeats itself, but is highly disadvantages if the patterns do not repeat. In the latter situation, the complex agents are more likely to fit noise, and cannot extrapolate well to the test data. We saw a precursor of this in the price prediction section, when the gradient boosting decision tree methods ended up overfitting the training data dramatically, and were even worse than logistic regression, which was still worse than even randomly guessing. 

Another point that we want to highlight is our choice of the time window. We choose the time window to be small enough so that we have more training instances. This is a big tradeoff that we have to balance because it is inherently possible that the agent just has a tough time making decision within such a short time window. There could be longer effects that don't manifest themselves on the scales we are searching, or ones that have a periodicity that escape from our narrow view. 

Overall, it appears as though we cannot say for sure whether our agents improve over one another or not, as all of their confidence intervals overlap. However, we can say that the approximate Q-learning framework allows the agents to capitalize on more information, and therefore at the very least it's a step in the right direction. Additionally, if we were to make a directional read, in 3/4 of the companies, our approximate learning agents outperform the baseline on average. In particular, in all four companies, our linear approximation agent outperforms the baseline more than 50\% of the time. This is a promising indication.

\section{Conclusion \& Future Work}
We have trained four agents and compared their performance on four technology companies' stocks. We can see that in certain situations,  our three RL agents were able to make greater profits than the baseline agent. This gives us hope that with further tuning, our RL agents could be useful in real life stock market prediction. 

In the future, we plan on extending our agents' state spaces in different ways. We could incorporate real financial features used by professionals such as market momentum, volume of stock traded and various indices of stock robustness that provide insight into the state of the market at any given time. This could potentially allow our agents to extract important information more easily from their state space. Another extension would be to add external data in the state of our agents. As mentioned before, information not available to our agent may directly influence the evolution of the prices of our stocks. As these are currently unknown to our agents, these evolutions may seem random to them. Extending our state space with new data could thus allow our agents to identify causes that directly influence the price evolutions within its state space. For example, we could potentially automatically parse the news related to the company we are interested in and try to spot events that are striking this company and which may have an impact on the evolution of its stocks.

A further improvement that we could make would be to revisit the price prediction and try more models that are geared for time series data. We could explore the use of LSTM's or other types of regression to identify upswings and downswings in the market. 

To summarize, we explored various methods for applying reinforcement learning to the stock market with the data we had available. The results were somewhat inconclusive, but there were promising indicators to show that our agents did outperform the baseline in certain situations. We made sure to rigorously and ethically evaluate our agents, and cleaned the data as best we could to ensure their success. We discussed what could have prevented the policies from converging, and also covered several areas for improvement. Identifying trends in the stock market is most definitely a difficult task. Our project elucidated how random the day to day motion of stock prices can be without factoring in external effects or long term trends. Perhaps one day, we can put all our faith into our algorithms. Until then, we shall put our faith instead in stock brokers. 

\appendix

\section{System Description}

Our code can be found at \url{https://github.com/TeamAI-2018/final_project} along with a README describing how to use the repository. A copy of the README is reproduced below for convenience.
\\\\
\textbf{HOW TO RUN AGENTS:}
\\\\
There are 4 agents in the repo that can be run. They reside in the following four directories:

\begin{itemize}
 \item Baseline\_agent: An agent that obeys simple logic to set a standard
 \item Q\_agent: An epsilon-greedy Q-learning agent
 \item linearAQ\_agent: An epsilon-greedy approximate Q-learning agent that uses linear function approximation
 \item NNQ\_agent: An epsilon-greedy approximate Q-learning agent that 
uses neural nets to perform function approximation
\end{itemize}

Within each of the directories are two important .py files. One will be a
variation of the agents name, containing a class that defines the agent. The other
is run.py, which is the only file the user needs to interact with. Simply type:
\begin{lstlisting}[language=Bash]
    $ python run.py
\end{lstlisting}
to run each agent. The Q-learning agent gives one the option to specify 
the input parameters. To see all the command line options with descriptions, type:
\begin{lstlisting}[language=Bash]
    $ python run.py -h
\end{lstlisting}
For the rest of the agents, in order to change the parameters, the user can simply edit the run.py file where the initializations are made.

\begin{figure}
    \centering
    \includegraphics[width=12cm]{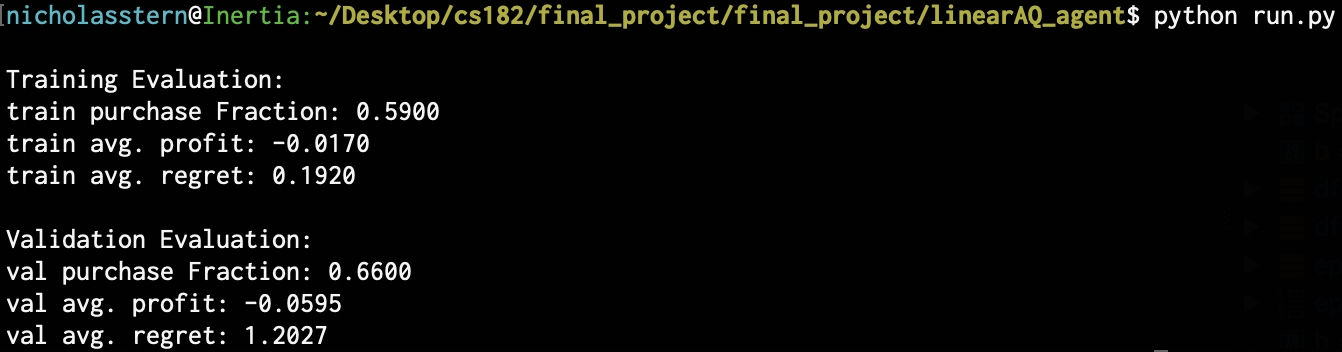}
    \caption{Example output after running "run.py"}
    \label{output}
\end{figure}

Each approximate Q learning agent will output the agent's performance on the training and validation sets. The metrics for evaluation are average profit, the average regret (see paper for in depth description), and the fraction of times the agent decide to buy. An example output looks like the output shown in figure \ref{output}.

Note that the exact Q-learning agent is not evaluated the same way by default, and will instead output an accuracy score for its decisions, and the fraction of its decisions that were purchases.
\\\\
\textbf{HOW TO EVALUATE AGENTS:}
\\\\
We have four companies' stock data, from Apple, Amazon, Google, Microsoft. To compare our four agents' performance on one particular company, open the file in the main directory called 'agent\_evaluation.py'. Locate the comment line that says '\# Parameter Initializations'. Under this line, change the company name to be the one you want to look at. For example, if you want to look at Microsoft's stock, you will type:

\begin{lstlisting}[language=Bash]
    company = "Microsoft"
\end{lstlisting}
Then in your command line, type:
\begin{lstlisting}[language=Bash]
    $ python agent_evaluation.py
\end{lstlisting}
After the code finishes running, you should be able to see a newly created '.csv' file, in the 'results' folder, that summarizes the performance of four agents. You also create a histogram graph in the 'images' folder.
\\\\
\textbf{OTHER CONTENTS:}
\\\\
Also in our Github repository, one can find price\_prediction.ipynb, where the price prediction code was written, alongside with EDA.ipynb, which was used to create Figure \ref{data_cut}. There is also a notebook in the Q\_agent folder that has plotting scripts for some hyperparameter tuning of that agent.

\section{Group Makeup}

We all designed our MDP and our agents together. We then separated the implementations within our group:
\begin{itemize}
    \item Ziyi (Queena) Zhou: Implemented the general Reinforcement Learning framework and the Deep Q-Learning agent.
    \item Nicholas Stern: Implemented the Q-Learning agent and the Q-Learning agent that uses linear function approximation.
    \item Julien Laasri: Implemented the price predictions, the Baseline agent and the evaluation of our agents.
\end{itemize}

\bibliographystyle{plain} 
\bibliography{citations}

\end{document}